\ifx\XeTeXversion\undefined\pdfoutput=1\fi
\documentclass[10pt,twocolumn,letterpaper]{article}

\usepackage[pagenumbers]{cvpr}

\usepackage{graphicx}
\usepackage{amsmath}
\usepackage{amssymb}
\usepackage{booktabs}
\usepackage{multirow}
\usepackage[numbers,sort,compress]{natbib}

\usepackage[pagebackref,breaklinks,colorlinks]{hyperref}
\usepackage[capitalize]{cleveref}
\crefname{section}{Sec.}{Secs.}
\Crefname{section}{Section}{Sections}
\Crefname{table}{Tab.}{Tabs.}
\crefname{table}{Tab.}{Tabs.}
\def\confName{Preprint}
\def\confYear{2026}

\newlength\savewidth

\def\cvprPaperID{0000}

\newcommand{\tool}[1]{\texttt{#1}}

\begin{document}
\raggedbottom
\setcounter{dbltopnumber}{1}
\setlength{\dbltextfloatsep}{12pt plus 2pt minus 2pt}
\setlength{\dblfloatsep}{10pt plus 2pt minus 2pt}

\title{Persistent Retrieval for Conversational Agents: A Study of Retrieval Efficiency}

\author{Karthik Ravula \qquad Sathwik Kowtha \qquad Rishabh Verma\\
{\tt\scriptsize karthik.ravula@servicenow.com \quad sathwik.kowtha@servicenow.com \quad rishabh.verma@servicenow.com}\\
Kala Mehak Jain \qquad Gunjit Mittal\\
{\tt\scriptsize kalamehak.jain@servicenow.com \quad gunjit.mittal@servicenow.com}\\
ServiceNow
}
\maketitle

\begin{abstract}
Conversational agents can retrieve information directly or delegate
retrieval to a subagent. This design choice affects how retrieval history
is maintained, what context each model call receives, and what information
must be re-retrieved in follow-up questions. We compare three designs:
skill-based retrieval in the main agent, a non-persistent retrieval
subagent that generates a findings response, and our persistent retrieval
subagent, which maintains retrieval history across turns and returns
retrieved evidence directly to the main agent. Our approach also allows
the main agent to request corrections from the same subagent.

We evaluate the three designs on enterprise questions posed independently,
within shared conversations, and as dependent follow-ups. Across five
multi-turn settings, our approach uses 43.3\% fewer recorded agent-model
tokens and incurs 45.0\% lower estimated API cost than skill-based
retrieval, while reducing average response time by 9.1\%. Compared with
non-persistent retrieval, it reduces response time by 15.6 seconds per
question, while using 65.1\% more tokens and incurring 29.1\% higher
estimated API cost.

Our approach shifts the retrieval subagent from producing synthesized
responses to maintaining retrieval state and supplying evidence that the
main agent can directly evaluate and use for answer generation.
\end{abstract}

\section{Introduction}
\label{sec:intro}

\begin{figure*}[t]
    \centering
  \includegraphics[width=0.80\textwidth]{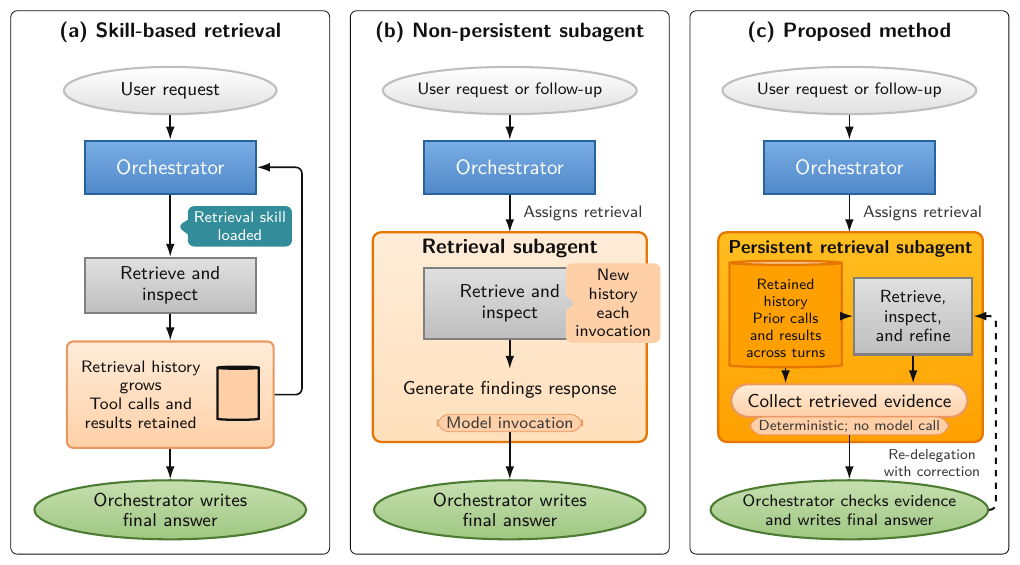}
  \caption{\textbf{Retrieval designs compared in our evaluation.}
  Skill keeps retrieval in the orchestrator. A non-persistent subagent
  generates a findings response. Our method retains retrieval history,
  returns evidence directly, and resumes the same session for corrections.}
    \label{fig:method}
\end{figure*}

Conversational agents often require multiple tool calls to answer a user
request. A harness can provide retrieval guidance as a \emph{skill} in the
main agent's context or delegate retrieval to a \emph{subagent} with its
own context~\citep{agentskills,claudesubagents}. This choice affects model
inputs and how retrieval history is maintained for later questions.

For enterprise users, an important practical consideration is whether an agent
can answer follow-up questions promptly without requiring previously
provided context to be repeated. With skill-based retrieval (\emph{Skill}), the main
agent, or orchestrator, uses a single context for both retrieval and
conversation. Consequently, results from earlier tool calls can
accumulate in the context and increase the input to subsequent model calls.

A non-persistent retrieval subagent starts each invocation with a new
history. It retrieves information and generates a findings response;
the orchestrator then generates the final answer. We call these two
response-generation stages \emph{double synthesis}. The findings response
can reduce the orchestrator's input, but the subagent does not retain its
retrieval history and may repeat earlier retrieval work.

Persistent and resumable agents carry context across
tasks~\citep{memgpt,claudesubagents}. Building on this capability, we
propose a persistent retrieval approach in which a subagent retains its
history across turns and returns evidence directly, without a separate
findings response. Here, \emph{evidence} means retrieval results
that the model uses to answer the retrieval task, together with the
queries or parameters used to obtain them.
Like ReWOO~\citep{rewoo}, our method separates retrieval
from answer generation, but keeps the retrieval session across turns so
the orchestrator can send corrections back with earlier queries and
results still available. For example, after retrieving a team's open
requests, the subagent can reuse that history when the user asks which
requests are overdue.

We ask two questions: (1) how do the three designs
trade response time against estimated API cost, and (2) how much context
does each design provide to the orchestrator? We evaluate independent
enterprise questions and conversations with follow-ups, measuring
response time, accuracy, estimated API cost, combined parent and child
token usage, and orchestrator input size. Appendix~\ref{ssec:related_work} discusses related
work.

\section{Methodology}
\label{sec:proposed_method}

Our method keeps retrieval work in a persistent subagent and returns
retrieved evidence directly to the orchestrator (\cref{fig:method}). The
orchestrator checks the evidence, requests corrections when needed, and
writes the final answer.
Appendix~\ref{app:retrieval_tools} describes our retrieval setup.

Within a conversation, the orchestrator and subagent maintain separate
contexts, each with its own budget. The subagent preserves its instructions,
messages, tool calls, and retrieved results across turns, so follow-up
questions can build on earlier retrieval. Before retrieving more, it checks
whether the information in its session is sufficient for the new question.
The orchestrator under Skill and our subagent use the same compaction engine:
large tool outputs are retained in workspace files and replaced in context
by compact previews. The orchestrator maintains its own context
independently and receives the subagent's output when it delegates retrieval.

In our implementation, we ask the subagent to briefly explain its retrieval
choices when invoking tools. When provided, these explanations accompany
the results and help the orchestrator assess whether the retrieval matches
the task. For overdue items, the explanation can identify the due-date and
open-status filters used by the query.

The subagent uses a Boolean argument, \tool{handoff\_after\_this}, to indicate
that the current retrieval step is expected to complete the task. When the
subagent sets this flag to true and the tool calls complete without errors,
the harness returns the evidence directly, without another model call to
generate a findings response. The orchestrator assesses the evidence
together with any accompanying explanation before writing the final answer.
If the evidence is incomplete or incorrect, it can redelegate retrieval with
additional guidance or context. The same subagent resumes from its existing
session, retaining its retrieval history while incorporating the additional
guidance. This separates retrieval from answer synthesis while preserving
the context needed for corrections.

\begin{table*}[!t]
  \centering
  \footnotesize
  \fontfamily{lmr}\selectfont
  \setlength{\tabcolsep}{5pt}
  \begin{tabular}{@{}llccrrrrr@{}}
    \toprule
    \textbf{Set} & \textbf{Method} & \textbf{Accuracy (\%)} & \textbf{Response time (s)} & \multicolumn{4}{c}{\textbf{Mean tokens per conversation}} & \textbf{Est. cost (\$)} \\
    \cmidrule(l){5-8}
    & & & & \textbf{Input} & \textbf{Cached} & \textbf{Uncached} & \textbf{Output} & \\
    \midrule
    \multirow{3}{*}{ITSM}
      & Non-persistent & 96.49 & 66.7 & 107.1k & 92.2k & 14.9k & 1.3k & 0.080 \\
      & Skill & 98.25 & \textbf{52.1} & 166.4k & 139.2k & 27.3k & 0.7k & 0.113 \\
      & Our method & 98.25 & 54.1 & 96.9k & 81.2k & 15.7k & 0.6k & \textbf{0.068} \\
    \addlinespace
    \multirow{3}{*}{HRSD}
      & Non-persistent & 93.75 & 78.3 & 108.0k & 93.8k & 14.2k & 1.4k & 0.080 \\
      & Skill & 95.83 & 52.9 & 160.7k & 134.5k & 26.2k & 0.8k & 0.111 \\
      & Our method & 93.75 & \textbf{51.9} & 100.4k & 87.7k & 12.7k & 0.5k & \textbf{0.061} \\
    \bottomrule
  \end{tabular}
  \caption{\textbf{Single-turn results.} Accuracy and response time use 57 ITSM
  and 48 HRSD questions per method (each single-turn conversation consists of
  one question); tokens are cumulative conversation means. Token values use
  thousands (k) and are rounded to one decimal place. Estimated API cost is in
  USD per conversation for recorded agent-model calls (\cref{ssec:metrics}).}
  \label{tab:quality_latency}
  \label{tab:token_study}
\end{table*}

\section{Experiments}
\label{sec:experiments}

We implement and evaluate all three approaches using ServiceNow's internal
agent harness with GPT-5.4 for agent model calls. All three methods share
a schema tool that looks up the metadata required for a question and a
Cypher tool that queries the records to answer it (Appendix~\ref{app:retrieval_tools}).

\subsection{Datasets}
\label{ssec:datasets}
The IT Service Management (ITSM) dataset consists of 57 questions covering
incidents, problems, changes, assignments, and service-level conditions.
The 48 HR Service Delivery (HRSD) questions cover employees, organizational
records, and HR cases. These questions are independent, even when asked
within the same conversation. In contrast, ITSM Complex comprises 69 questions with multi-turn dependencies: later questions may
refer to information retrieved or discussed in earlier turns.
\Cref{tab:dataset_samples} shows five sample questions from each dataset.

For ITSM and HRSD, the single-turn study poses questions independently
to each method. The multi-turn study uses two settings: $\leq20$ groups
questions into conversations of up to 20 questions, while \emph{Full}
places the entire question set in one conversation. ITSM Complex uses
three conversations of 19, 19, and 31 questions
(\cref{app:evaluation_protocol}).

\begin{table*}[!t]
  \centering
  \footnotesize
  \fontfamily{lmr}\selectfont
  \setlength{\tabcolsep}{5pt}
  \begin{tabular}{@{}llccrrrrr@{}}
    \toprule
    \textbf{Set} & \textbf{Method} & \textbf{Accuracy (\%)} & \textbf{Response time (s)} & \multicolumn{4}{c}{\textbf{Mean tokens per question}} & \textbf{Est. cost (\$)} \\
    \cmidrule(l){5-8}
    & & & & \textbf{Input} & \textbf{Cached} & \textbf{Uncached} & \textbf{Output} & \\
    \midrule
    \multirow{3}{*}{ITSM ($\leq20$)}
      & Non-persistent & 94.74 & 73.0 & 142.8k & 115.0k & 27.8k & 1.5k & \textbf{0.120} \\
      & Skill & 96.49 & 68.2 & 461.8k & 379.6k & 82.1k & 0.6k & 0.309 \\
      & Our method & 96.49 & \textbf{66.4} & 246.5k & 188.0k & 58.4k & 0.5k & 0.201 \\
    \addlinespace
    \multirow{3}{*}{HRSD ($\leq20$)}
      & Non-persistent & 89.58 & 71.1 & 147.3k & 126.1k & 21.2k & 1.4k & \textbf{0.105} \\
      & Skill & 91.67 & \textbf{60.3} & 264.6k & 238.7k & 25.9k & 0.8k & 0.136 \\
      & Our method & 91.67 & 63.5 & 218.8k & 179.6k & 39.2k & 0.5k & 0.150 \\
    \midrule
    \multirow{3}{*}{ITSM (Full)}
      & Non-persistent & 94.74 & 80.3 & 256.9k & 198.5k & 58.4k & 1.4k & \textbf{0.216} \\
      & Skill & 96.49 & 74.0 & 766.8k & 705.3k & 61.5k & 0.7k & 0.510 \\
      & Our method & 96.49 & \textbf{63.2} & 430.3k & 352.3k & 78.0k & 0.5k & 0.290 \\
    \addlinespace
    \multirow{3}{*}{HRSD (Full)}
      & Non-persistent & 91.67 & 81.3 & 178.6k & 131.8k & 46.8k & 1.4k & \textbf{0.170} \\
      & Skill & 87.50 & 81.1 & 669.9k & 622.9k & 47.0k & 0.9k & 0.365 \\
      & Our method & 89.58 & \textbf{63.1} & 357.8k & 306.9k & 51.0k & 0.4k & 0.210 \\
    \midrule
    \multirow{3}{*}{ITSM Complex}
      & Non-persistent & 91.30 & 79.6 & 148.0k & 115.0k & 33.0k & 1.4k & 0.133 \\
      & Skill & 91.30 & 58.3 & 401.5k & 278.0k & 123.5k & 0.6k & 0.387 \\
      & Our method & 89.85 & \textbf{54.3} & 210.8k & 183.8k & 27.0k & 0.5k & \textbf{0.121} \\
    \bottomrule
  \end{tabular}
  \caption{\textbf{Multi-turn results.} ITSM and HRSD use 57 and 48 questions
  per setting; ITSM Complex uses 69 (\cref{app:multiturn_sessions}).
  Tokens (k) sum parent and child calls per question, including cached input.
  Estimated API cost is in USD per question for recorded agent-model calls
  (\cref{ssec:metrics}).}
  \label{tab:multiturn_results}
  \label{tab:multiturn_tokens}
  \label{tab:complex_results}
\end{table*}

\subsection{Methods and baselines}
\label{ssec:baselines}
With skill-based retrieval, the orchestrator invokes retrieval tools
directly and keeps the retrieval skill and tool results in its own context.

A non-persistent retrieval subagent starts with a new message history
for each invocation, retrieves the data, and
generates a findings response. The orchestrator receives that response and writes the answer.

Our method uses a subagent that retains retrieval history across turns
and returns evidence directly to the orchestrator
(\cref{sec:proposed_method}).

\subsection{Evaluation metrics}
\label{ssec:metrics}
We use ServiceNow's internal test-automation framework, where GPT-5.4-mini
serves as the LLM judge, comparing the orchestrator's response with the
ground-truth answer to assign a quality score and a consistency verdict.
We count a question as a pass when the score is greater than zero and the
judge marks the response as consistent with the ground truth. We calculate
accuracy for each run and then average the three percentages.

Response time is the benchmark-recorded time to answer one question,
reported in seconds. We count each recorded parent and child model call
once, summing usage per conversation for the single-turn study
and per question for multi-turn. Cached input is part of input, uncached
input is the remainder, and total tokens are input plus output.
\Cref{app:measurement_details} gives the accounting equation.
\Cref{tab:quality_latency,tab:multiturn_results} report
the mean of three independent runs for each method and setting.

We estimate API cost using OpenAI's Standard GPT-5.4 prices: \$2.50 per
million uncached input tokens, \$0.25 per million cached input tokens,
and \$15 per million output tokens~\citep{openai_gpt54_pricing}. These
rates rise to \$5, \$0.50, and \$22.50 for calls exceeding 272k input
tokens, including cached input. We price all tokens in each call at its
applicable rates before averaging. Estimates cover the recorded agent-model
calls, not actual deployment bills; retrieval-service and judge calls,
infrastructure, and negotiated or regional pricing adjustments are excluded.

\subsection{Results}
\label{ssec:results}

Our method answers faster than non-persistent retrieval
in every evaluated setting. Across multi-turn settings, it also has lower
average response time, token usage, and estimated API cost than Skill.

Our method achieves the lowest response time and estimated cost on ITSM Complex,
which tests dependent follow-ups (\cref{tab:complex_results}). It takes
54.3 seconds per question, versus 79.6 for non-persistent retrieval and
58.3 for Skill, and costs \$0.121, versus \$0.133 and \$0.387, respectively.
Fewer uncached-input and output tokens explain the lower cost despite
41.4\% more total tokens than non-persistent retrieval. Accuracy is 89.85\%,
versus 91.30\% for both baselines.

The ITSM Complex ablation (Appendix~\ref{ssec:component_ablation},
\cref{tab:component_ablation}) shows a larger response-time and cost
increase without direct return (17.7 seconds, 42\% higher cost) than
without persistence (4.3 seconds, 8\% higher cost). Reported
accuracy remains 89.85\% in all three configurations.

Across all five multi-turn settings, our method has the lowest
question-weighted average response time (\cref{tab:multiturn_results}).
Compared with Skill, it is 9.1\% faster, uses 43.3\% fewer total tokens,
and reduces estimated API cost by \$0.157 (45.0\%) per question; only on
HRSD ($\leq20$) is Skill faster (60.3\,s vs.\ 63.5\,s) and lower-cost
(\$0.136 vs.\ \$0.150). Against non-persistent retrieval, our method answers
15.6 seconds faster but uses 65.1\% more tokens and costs \$0.043 (29.1\%)
more per question on average (\cref{fig:benchmark_time_cost}). Retained
subagent history adds input across turns, helping explain the token-usage
trade-off.

In the analyzed multi-turn traces, our method re-delegated 11 of 279
questions (3.9\%) to obtain missing evidence or recheck query filters.
The reported token and cost totals include these calls.

In single-turn settings, our method has response times close to Skill,
with the fewest tokens and lowest estimated cost on both datasets
(\cref{tab:quality_latency}). Non-persistent retrieval is about
12.6 seconds slower on ITSM and 26.4 on HRSD. Across three
runs of each setting, the number of questions answered correctly varied
by about two or three; our method and either baseline differed by no more
than one.

Our method's mean peak orchestrator inputs are smaller than with Skill
but larger than with non-persistent retrieval across all five settings
(\cref{tab:parent_context}). Peak input is the largest single-call input
per conversation, not cumulative usage.

\medskip
\noindent\begin{minipage}{\columnwidth}
  \centering
  \footnotesize
  \setlength{\tabcolsep}{5pt}
  \begin{tabular}{@{}cccc@{}}
    \toprule
    \textbf{Set} & \textbf{Non-persistent} & \textbf{Skill} & \textbf{Our method} \\
    \midrule
    ITSM ($\leq20$) & 66.5k & 180.0k & 102.4k \\
    HRSD ($\leq20$) & 60.6k & 122.4k & 95.0k \\
    \midrule
    ITSM (Full) & 155.4k & 493.8k & 242.3k \\
    HRSD (Full) & 106.4k & 335.5k & 192.3k \\
    \midrule
    ITSM Complex & 68.8k & 188.6k & 104.0k \\
    \bottomrule
  \end{tabular}
  \captionof{table}{Mean of each conversation's largest single-call
  orchestrator input, in thousands of tokens (k). ITSM and HRSD use three
  conversations per shorter run and one per full-length run; ITSM Complex
  uses three.}
  \label{tab:parent_context}
\end{minipage}

\medskip
\looseness=-1
Our method reduces response-generation time by 12.3--13.6 seconds per
question versus non-persistent retrieval (\cref{tab:synthesis_time}),
avoiding intermediate findings generation. Timings cover matched
single-turn subsets and multi-turn ITSM Complex, excluding retrieval
and other overhead; they are not end-to-end savings.

\medskip
\noindent\begin{minipage}{\columnwidth}
  \centering
  \footnotesize
  \fontfamily{lmr}\selectfont
  \setlength{\tabcolsep}{4pt}
  \begin{tabular}{@{}cccc@{}}
    \toprule
    \textbf{Set} & \textbf{Non-persistent} & \textbf{Our method} & \textbf{Time saved} \\
    \midrule
    ITSM & 19.58 & 7.30 & 12.28 \\
    HRSD & 19.02 & 5.44 & 13.58 \\
    \midrule
    ITSM Complex & 19.66 & 6.96 & 12.70 \\
    \bottomrule
  \end{tabular}\par
  \captionof{table}{Mean model-call time (s) per question for answer
  generation. ITSM (52) and HRSD (32) use matched single-turn questions
  where both methods delegated and answered. ITSM Complex includes all 69
  multi-turn questions, with or without delegation.}
  \label{tab:synthesis_time}
\end{minipage}

\section{Discussion and Conclusions}
\label{sec:discussion}

In this paper, we introduce a persistent retrieval subagent that returns
evidence directly to the orchestrator. Direct evidence return avoids
intermediate findings generation, while persistence
preserves retrieval context for follow-ups and corrections. The ablation
in Appendix~\ref{ssec:component_ablation} shows larger response-time and cost
increases without direct return than without persistence. The orchestrator
evaluates the evidence, requests targeted corrections when needed, and
writes the final answer. Across the main evaluations, our method maintains comparable
accuracy while reducing average multi-turn response time, tokens, and
estimated cost relative to Skill. Against non-persistent retrieval, it
responds faster in all seven settings, but uses more tokens and costs more
on average in multi-turn conversations.

\section*{Limitations}
\label{sec:limitations}
The subagent sets the completion flag before seeing the retrieval result.
A tool can succeed while returning incomplete or irrelevant evidence,
which reaches the orchestrator without another child-model check. The
correction mechanism in \cref{sec:proposed_method} depends on the
orchestrator recognizing the gap; it may instead answer without requesting
further evidence.

Our evaluation focuses on structured enterprise records in a demonstration
environment. The results show how the methods performed on these question
sets; understanding how they extend to other tasks will require broader
testing and a breakdown by question difficulty.
Results may also vary with the agent harness and its configuration.
\label{sec:main_text_end}

{\small
\bibliographystyle{abbrvnat}
\bibliography{egbib}
}

\clearpage
\appendix
\section*{Appendix}
\section{Evaluation Protocol}
\label{app:evaluation_protocol}

\subsection{Single-turn question sets}
Each method answers 57 ITSM and 48 HRSD questions posed independently.
The questions cover the domains described in \cref{ssec:datasets}, with
retrieval work ranging from finding an attribute to following several
related records.

\subsection{Multi-turn session construction}
\label{app:multiturn_sessions}
We use two conversation settings. In the shorter setting, the 57 ITSM
questions are divided into conversations of 20, 20, and 17 questions;
the 48 HRSD questions form conversations of 20, 20, and 8. In the
full-length setting, each dataset remains in one conversation. Each
question is asked once in its conversation, as a separate user turn.
History accumulates within a
conversation and is reset at the next conversation boundary. Under
our method, retrieval tasks within the same conversation resume the child's
existing session.

The separate ITSM Complex evaluation uses 69 questions per method,
grouped into conversations of 19, 19, and 31 questions. It includes
follow-up questions that depend on earlier turns in the conversation.

All three methods receive the same question sequences, including standalone
questions and follow-ups. Shared history makes outcomes
within a conversation dependent, and fixed question order couples
conversation depth with question difficulty. We do not treat turns as
independent trials.

\section{Measurement Details}
\label{app:measurement_details}
\setlength{\intextsep}{6pt}

For single-turn results (\cref{tab:quality_latency}), let $\mathcal{L}_i$
contain the model calls in conversation $i$. Mean usage over $N$
conversations is
\begin{equation}
  \overline{X}=\frac{1}{N}\sum_{i=1}^{N}\sum_{j\in\mathcal{L}_i}X_{ij},
  \quad X\in\{I,C,O\},
  \label{eq:token_accounting}
\end{equation}
where $I$, $C$, and $O$ denote input, cached-input, and output tokens.
Mean uncached input is $\overline{U}=\overline{I}-\overline{C}$.

For call $j\in\mathcal{L}_i$, let $U_{ij}=I_{ij}-C_{ij}$ and let
$r_{u}(I_{ij})$, $r_{c}(I_{ij})$, and $r_{o}(I_{ij})$ denote uncached-input,
cached-input, and output prices in USD per token. Estimated cost is
\begin{equation}
  \mathrm{Cost}_{ij} = U_{ij}\,r_{u}(I_{ij}) + C_{ij}\,r_{c}(I_{ij}) + O_{ij}\,r_{o}(I_{ij}),
  \label{eq:cost_accounting}
\end{equation}
where $(r_{u}, r_{c}, r_{o})(I) = (2.5, 0.25, 15.0)\times 10^{-6}$ for
$I \le 272\text{k}$ and $(5.0, 0.50, 22.5)\times 10^{-6}$ for
$I > 272\text{k}$ (\cref{ssec:metrics}). Across all methods and datasets,
we sum call costs per conversation or question and average these totals, as
for token usage.

We count each orchestrator and subagent call once by task identifier.
Input includes cached tokens; output includes tool-call arguments,
intermediate responses, and final answers. Rounding follows aggregation.

For multi-turn results (\cref{tab:multiturn_results}), we sum call usage
per question and average over 57 ITSM, 48 HRSD, or 69 ITSM Complex questions.
Combined results weight questions equally across the five settings;
repeated runs within each method and setting also receive equal weight.

Totals include calls finishing after answer display and may therefore
extend beyond the response timer. Separately recorded retrieval-service
model calls are excluded.

\Cref{tab:parent_context} averages each conversation's largest single-call
orchestrator input within each condition, weighting runs equally. Inputs
include instructions, tool definitions, and conversation history.

\subsection{Response-generation timing}

For \cref{tab:synthesis_time}, we sum subagent findings-response and
orchestrator final-answer call durations per question, then average over
questions. Direct-return paths omit child findings generation; any
recorded child completions remain included in the totals.

ITSM and HRSD use 52 and 32 matched single-turn questions where both methods
delegated and answered; ITSM Complex uses all 69 multi-turn questions.
Time saved is the non-persistent mean minus ours. Retrieval and other
overhead are excluded, unlike the end-to-end times in
\cref{tab:quality_latency,tab:multiturn_results}.

\begin{table*}[!t]
  \centering
  \footnotesize
  \fontfamily{lmr}\selectfont
  \setlength{\tabcolsep}{5pt}
  \begin{tabular*}{\textwidth}{@{\extracolsep{\fill}}lrrrrrrr@{}}
    \toprule
    \textbf{Configuration} & \shortstack{\textbf{Accuracy}\\\textbf{(\%)}} &
    \shortstack{\textbf{Response}\\\textbf{time (s)}} &
    \textbf{Total tokens} & \shortstack{\textbf{Uncached}\\\textbf{input}} &
    \shortstack{\textbf{Schema}\\\textbf{calls}} &
    \shortstack{\textbf{Cypher}\\\textbf{calls}} & \textbf{Est. cost (\$)} \\
    \midrule
    No persistence & 89.85 & 58.6 & \textbf{191.8k} & 32.0k & 49 & 71 & 0.1302 \\
    No direct return & 89.85 & 72.0 & 302.1k & 35.4k & 31 & 80 & 0.1712 \\
    Our method & 89.85 & \textbf{54.3} & 211.3k & \textbf{27.0k} &
    \textbf{23} & \textbf{59} & \textbf{0.1205} \\
    \bottomrule
  \end{tabular*}
  \caption{\textbf{Component ablation.} Persistent retrieval state and
  direct evidence return on ITSM Complex. Call counts are averaged across
  runs and rounded to the nearest integer;
  time, tokens, and cost are per-question means. Tokens use thousands (k).}
  \label{tab:component_ablation}
\end{table*}

\section{Retrieval Tools and Context Reuse}
\label{app:retrieval_tools}
The methods use a schema tool to retrieve the metadata required for a
question (tables, fields, and relationships) and, following query rules in
the prompt, write and run a Cypher query over that schema to retrieve the
records needed to answer it. Skill retains retrieval outputs in the
orchestrator. A non-persistent subagent receives orchestrator-provided
context but starts each invocation with fresh child history. Our method
retains metadata, queries, and results across turns, allowing earlier
evidence to be reused alongside new retrieval.

For this trace analysis, we use one ITSM Complex run per method
(69 questions); \cref{tab:component_ablation} instead averages ablation runs.
Our method made 23 metadata lookups and 59 Cypher queries, compared
with 36 lookups and 85 queries for Skill and 64 lookups and 98 queries
for non-persistent retrieval. Thus, our method performed less
recorded retrieval work than either baseline in these runs.

On ITSM Complex, mean response-generation call time was 6.96 seconds per
question for our method, compared with 19.66 seconds for non-persistent
retrieval---a difference of 12.70 seconds (\cref{tab:synthesis_time}).
Both measurements include all recorded child and orchestrator
response-generation calls. These timings exclude retrieval and other
overhead and are not end-to-end savings.

Across one analyzed run for each of five multi-turn settings, 249 of 274
subagent invocations returned evidence directly through the completion
flag (\cref{sec:proposed_method}), with no errors in the marked tool calls.
The remaining 25 ended with explicit tool-output selection (14), text (8),
or an empty model response (3). These counts measure direct-return use;
answer accuracy is reported separately.

Together, these observations show that our method exhibits both
lower recorded retrieval work and lower response-generation time than
non-persistent retrieval.

\section{Related Work}
\label{ssec:related_work}

Prior work examines how agents retrieve information and how they plan
and coordinate retrieval tasks. Retrieval-augmented generation supplies
retrieved passages, while Toolformer and ReAct use tools to obtain
information~\citep{rag,toolformer,react}. Skills and multi-agent harnesses
organize this work through reusable guidance or collaborating
agents~\citep{agentskills,autogen}. ReWOO separates planning from execution
and passes collected evidence to a Solver; LLMCompiler schedules dependent
function calls, including parallel execution and
replanning~\citep{rewoo,llmcompiler}. Like ReWOO, our method separates
retrieval from answer generation, but keeps the retrieval session across
turns so the orchestrator can send corrections back with earlier queries
and results still available. The harness collects selected tool outputs
and their arguments unchanged, without another model call to write a
findings response.

MemGPT and resumed subagents preserve context across
tasks~\citep{memgpt,claudesubagents}. Self-Refine and Reflexion use feedback or
retained reflections to improve later attempts~\citep{selfrefine,reflexion}.

\section{Ablation Study}
\label{ssec:component_ablation}
Our approach uses two mechanisms to improve the efficiency of multi-step
retrieval: persistent retrieval state and direct evidence return. Because
both are enabled in the full system, the main evaluation does not reveal
how much each contributes individually. We therefore perform an ablation
on ITSM Complex to examine whether the efficiency gains come primarily
from reusing information across questions, avoiding intermediate
findings-generation, or their combination. We disable each mechanism while
retaining the other and compare response time, token usage, and estimated
cost (\cref{tab:component_ablation}). We ran each ablation configuration
three times. The full-method reference uses earlier measurements, with accuracy from
\cref{tab:multiturn_results}.

The observed execution patterns suggest complementary efficiency benefits.
In this comparison, the full method achieves the lowest response time and
estimated cost, while requiring fewer uncached input tokens and fewer
schema and Cypher calls. Persistent state allows schemas and results from
earlier questions to be reused, reducing the need for repeated retrieval.
Without direct evidence return, the subagent instead generates a findings
response for the orchestrator, adding generation overhead.

The ablation also reveals a token-usage trade-off. Disabling persistence
produces the fewest total tokens because the subagent no longer carries
its retrieval history across questions. Persistence can reduce retrieval
work at the cost of a larger context, while direct evidence return reduces
intermediate generation. Together, the mechanisms combine retrieval reuse
with a shorter path from retrieval to orchestration. Their combination
gives the best response-time and estimated-cost profile in this comparison,
rather than minimizing every individual metric.

\section{Acknowledgements}
\label{app:acknowledgements}

We thank Malhar Chaudhari and Rajagopalan Govind for reviewing this paper
and providing valuable feedback.

\onecolumn
\section{Sample Questions}
\label{app:sample_questions}

\medskip
\begingroup
\setlength{\fboxsep}{10pt}
\setlength{\fboxrule}{0.3pt}
\newcommand{\samplequestions}[2]{%
  \noindent\fbox{%
    \begin{minipage}{\dimexpr\linewidth-2\fboxsep-2\fboxrule\relax}
      \raggedright
      \textbf{#1}
      \begin{list}{\arabic{enumi}.}{%
        \usecounter{enumi}
        \setlength{\leftmargin}{1.8em}
        \setlength{\labelwidth}{1.2em}
        \setlength{\labelsep}{0.6em}
        \setlength{\topsep}{6pt}
        \setlength{\itemsep}{5pt}
        \setlength{\parsep}{0pt}
      }
        #2
      \end{list}
    \end{minipage}%
  }\par
}

\noindent\begin{minipage}{\columnwidth}
\captionof{table}{\textbf{Sample questions from the evaluated datasets.} The
datasets cover filtering, aggregation, relationship traversal, and
follow-up retrieval. The ITSM Complex examples are five consecutive
turns from one conversation.}
\label{tab:dataset_samples}
\medskip

\samplequestions{ITSM}{
  \item Get incident number and short description for incidents with
    contact type email and category inquiry.
  \item Display incidents that are older than 3 days and have not been
    resolved yet.
  \item Total incident reassignments by assigned user for priority 1
    incidents, ordered by count.
  \item What is the workaround for incident number INC0010004?
  \item What are the incidents that were caused by Change CHG0000007?
}
\end{minipage}\par

\bigskip
\samplequestions{HRSD}{
  \item Who are the contacts for the building ``Adelas Building''?
  \item Which departments use the cost center with the account number
    ACN00001?
  \item What is the job title of the job profile with the job code 1001?
  \item What is the name of the company that is the parent company of
    the company located in Shanghai?
  \item What is the name of the business unit associated with the
    department that uses the cost center with the code 123?
}

\bigskip
\samplequestions{ITSM Complex}{
  \item Who reported INC1108477?
  \item Which team is currently assigned to INC1108477?
  \item What service and service offering does INC1108477 belong to?
  \item When did the outage begin and when was it fully restored?
  \item How long did the outage last from start to resolution?
}
\endgroup

\clearpage
\section{Benchmark Comparisons}
\label{app:benchmark_comparisons}

\noindent\begin{minipage}[t]{0.485\textwidth}
  \centering
  \includegraphics[page=1,width=\linewidth]{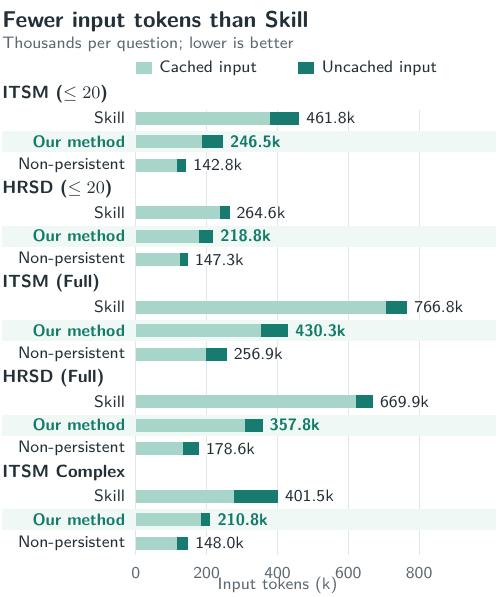}
  \captionof{figure}{Input tokens across all five multi-turn settings. Stacks
  show cached and uncached input; labels give input tokens per
  question in thousands (k) (\cref{tab:multiturn_results}).}
  \label{fig:benchmark_tokens}
\end{minipage}\hfill
\begin{minipage}[t]{0.485\textwidth}
  \centering
  \includegraphics[page=2,width=\linewidth]{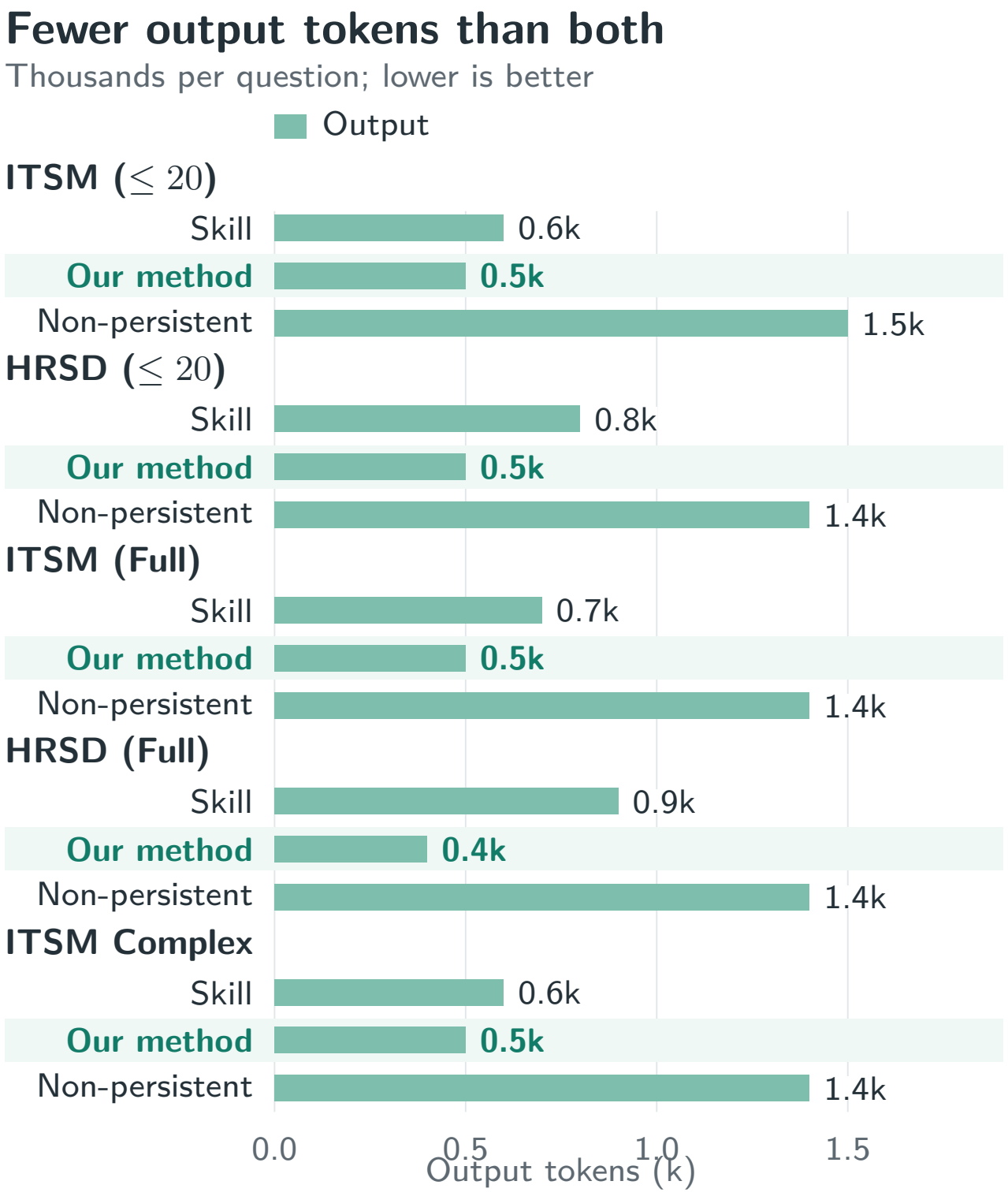}
  \captionof{figure}{Output tokens across the same five settings. Bars and
  labels show output tokens per question in thousands (k) (\cref{tab:multiturn_results}).}
  \label{fig:benchmark_output}
\end{minipage}

\bigskip
\noindent\begin{minipage}{\textwidth}
  \centering
  \includegraphics[page=3,width=\textwidth]{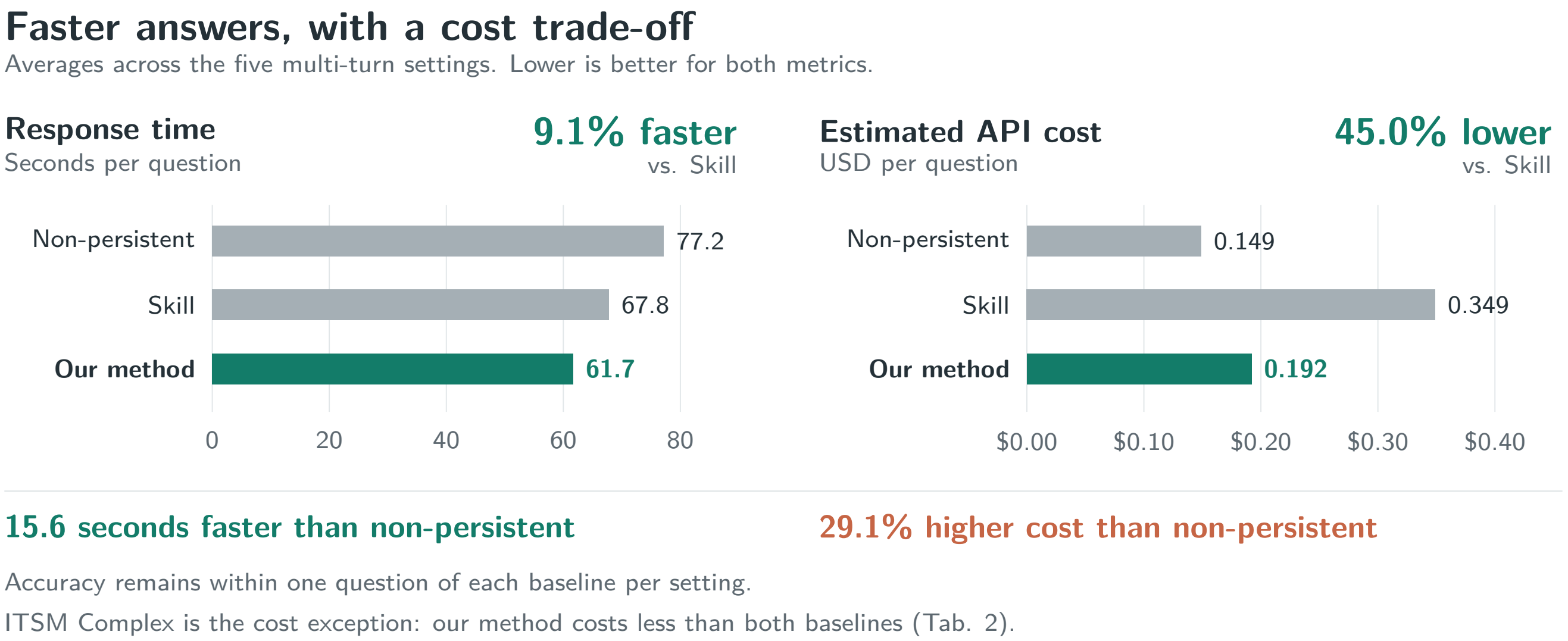}
  \captionof{figure}{\textbf{Overall speed and estimated-cost trade-off.}
  Question-weighted averages across the five settings. Costs cover recorded
  agent-model calls only (\cref{ssec:metrics}), not deployment bills.
  Per-setting accuracy and other results are in \cref{tab:multiturn_results}.}
  \label{fig:benchmark_time_cost}
\end{minipage}

\end{document}